# A 3D Segmentation Method for Retinal Optical Coherence Tomography Volume Data

Yankui Sun, Tian Zhang

*Abstract*—With the introduction of spectral-domain optical coherence tomography (OCT), much larger image datasets are routinely acquired compared to what was possible using the previous generation of time-domain OCT. Thus, the need for 3-D segmentation methods for processing such data is becoming increasingly important. We present a new 3D segmentation method for retinal OCT volume data, which generates an enhanced volume data by using pixel intensity, boundary position information, intensity changes on both sides of the border simultaneously, and preliminary discrete boundary points are found from all A-Scans and then the smoothed boundary surface can be obtained after removing a small quantity of error points. Our experiments show that this method is efficient, accurate and robust.

## Introduction

THE retina is a complex organization composed of a transparent layer of tissue, three-dimensional automatic segmentation algorithm developed for frequency-domain OCT retinal images has significant value for automatic diagnosis of eye diseases such as glaucoma, diabetic retinopathy, etc.

Because the significance of a three-dimensional automatic segmentation of frequency-domain OCT retinal images [1,2], the development of a rapid, accurate and reliable 3D segmentation algorithms, which are different from the common two-dimensional image segmentation, becomes necessary. However, OCT images have some unique characteristics, which make three-dimensional segmentation difficult. For examples, there are serious speckle noises prevalent in OCT images [3], reducing the image quality, especially the sharpness of the edges; and the retina is at the back of the eye, reducing the OCT signal intensity. To solve these problems, researchers across the world have made many achievements, developed a variety of different three-dimensional segmentation algorithms.

Mona et al. proposed a graph-search based three-dimensional OCT retinal image segmentation algorithm [4]. Zawadzki et al. proposed a segmentation method with support vector machine (SVM) [5]. Fernandez et al. segmented the OCT retinal images by calculating the coherence matrix to replace the original pixel value as the stratification criteria [6]. Mujat et al.. used spline deformable surface model to calculate the retinal location of ILM, the top boundary [7]. These methods above mentioned share the similar disadvantages that they all require lots of calculation, thus lowering their efficiency, and they are all sensitive to noises in OCT images, so they don't enjoy high robustness which is required for commercial clinical application.

In a different strategy, Fabritius designed a simple approach [8], which currently can only mark the outer boundary of the retina, i.e. ILM (internal limiting membrane) and RPE (retinal pigment epithelium). Its basic idea is to find a preliminary position for ILM and RPE in one-dimensional A-scan based on two distinct characteristics of the border, then , to use stepwise refinement, successively narrow the search range, and continue to rule out obvious wrong boundary points, and ultimately get reasonably accurate positioning of the boundaries. The one major advantage of this method is it needs less calculation (than those methods mentioned above). But its disadvantage is that it is sensitive to noises, for its looking for RPE boundary is based on the assumed fact that this border region is the brightest in the retinal OCT image, but in images with serious noises this is not always true; this method requires much iteration, which is very time-consuming; at the same time, the method can only find ILM and RPE boundaries. In this paper, we propose a new 3D segmentation method by using pixel intensity, boundary position information, intensity changes on both sides of the border simultaneously, which can segment three important tissue surfaces automatically, efficiently and robustly.

## Algorithm Description

*The basic principle*

It is discovered that the most significant features of the boundaries of retina usually are not the intensity of the pixels as utilized by Fabritius, etc., but the intensity changes on both sides of the border. For example, in OCT retina images the RPE boundary region is usually one of the brightest regions, but the intensity change between the both sides of RPE is more significant. In [8], this important information has not been applied. So by designing and applying a proper three-dimensional convolution operator on the 3D retinal OCT volume data, we can get a new volume data which intensify the pixel change across important boundaries such as ILM and RPE. This new volume data can be used as a better basis for finding the boundaries.

Meanwhile, Fabritius' algorithm uses pixel intensity and other indicators to determine the initial location of discrete boundary points, and then uses the already-known

*Research supported by National Natural Science Foundation of China (No. 60971006).

Yankui Sun is with the Department of Computer Science and Technology, Tsinghua University, China (phone: 86-10-62782609; fax: 86-10-62771138; e-mail: syk@ mail.tsinghua.edu.cn).

Tian Zhang is with the Department of Computer Science and Technology, Tsinghua University, China. (email: kasperlzhang@gmail.com)

approximate position of the boundaries to eliminate false boundary points. This strategy is not the most efficient and optimum use of the boundary position information. In fact, in the first attempt to determine the required discrete boundary points in each A-scan, the z coordinates should already be taken into account. Using this improved strategy, in the first step to get the coordinates of a large number of discrete points, we can basically rule out the obvious error boundary points, making the step-by-step refinement method unnecessary.

*Algorithm Flow*

Based on the above analysis, we designed a new three-dimensional segmentation method of the retina, which at the present can determine ILM, RPE and IS/OS (inner segment-outer segment) boundaries. The basic idea of this method is, based on the characteristics of an interested boundary, to design an operator and apply it on the original volume data so that the pixel value on the desired boundary in the new volume data is likely to be the maximum value in its A-scan, making the new volume data a better indication to find the desired boundaries, thereby reducing or even eliminating the necessity of refinement process. Taking the RPE boundary as example, it is characterized by the fact that its pixel intensity is the brightest of all in every A-scan [8], the brightness of the points above this boundary is significantly greater than the points below it, and that it is located on the bottom of retinal OCT images. Based on these characteristics, the method to determine the location of the boundary RPE is as follows:

**Step 1** Design and apply a 3D differential filter on original volume data $V$ to get a volume data $D$;

**Step 2** Perform a three-dimensional smoothing procedure on $V$ to get a volume $S$;

**Step 3** Generate a new volume data, denoted as $I$, to further enhance the RPE boundary by using position information.

$$I_{i,j,k} = k*(D_{i,j,k} + S_{i,j,k})$$

where $k$ is the z coordinate of each pixel;

**Step 4** In each A-scan of $I$, take the point which has the largest pixel value as the preliminary location of RPE boundary.

As mentioned above analysis, the largest pixel value is very likely to lie on the RPE boundary positions, so the results obtained in most cases is a good estimate of the location of RPE.

**Step 5** Get a reliable RPE boundary in 3D volume data by simple smoothing on the results obtained in Step 4.

The algorithm flow chart is shown in Fig. 1.

When the boundary RPE is determined, the boundaries IS/OS and ILM can be obtained in a similar way. The boundary IS/OS can be obtained by using a different 3D differential filter on a truncated volume data which is the one abandoning all the pixels of the boundary RPE and those below it from the original volume data $V$. Further, ILM can be detected at the same way.

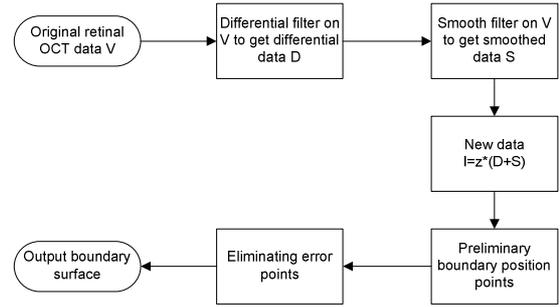

Figure 1. Flowchart of the determination of one boundary

Comparing with Fabritius' method in [8], one of the most important innovations of our method is using the boundary location information to get a better approximation of boundary points, hence eliminating the necessity of stepwise refinement; thereby greatly reducing calculation cost and gets better result at the same time. In addition, our method can segment more boundary surfaces.

EXPERIMENTAL RESULT

The experimental data used in our experiments consist of two sets of 480 × 300 × 99 3D retinal OCT volume data, one normal and the other abnormal. These data were obtained by the 3D spectrum domain optical coherence tomography (OCT) system based on fiber-based Michelson interferometer [9]. The measurable depth of the OCT imaging is up to 3.4 mm for the detection part is of high speed spectrometer with 0.05 mm spectrum resolution. The signal-noise-ratio(SNR) and axial resolution of this OCT system are 51 dB and 8.5 mm, respectively. As mentioned before, these data are polluted by severe speckle noises, have relatively low resolution, and in the abnormal data set, there are many abnormalities, further increasing the difficulty of segmentation. Fig.2 is one slice in the normal data set, and Fig.3 is one in the abnormal set. The images were processed on Intel Pentium Dual-Core E5200 2.50GHz, 2G RAM, Nvidia Geforce 9600GT's personal computer, using code developed by MATLAB 2007. In our experiments conducted on a 480 × 300 × 99 OCT retinal volume data, each boundary can be determined within 5 seconds.

Fig.4 and Fig.5 demonstrates the discrete boundary points and ideal boundary of RPE in Fig. 2, which are the processing results after step 4 and step 5 respectively. The processing results with three boundaries ILM, RPE and IS/OS are shown in Fig. 6. Using the ILM and RPE boundary locations, we can calculate the total retinal thickness maps, as shown in Fig.7. Retinal thickness is very indicative in the diagnosis of many retinal diseases, such as glaucoma, diabetic retinal edema, etc.

It is also possible to visualize the position of one boundary surface fetched by our method in 3D spaces, to have a better grasp on their spatial characteristics, as shown in Fig. 8.

Furthermore, our method is also able to deal with abnormal data obtained on patients, as show in Fig. 9.

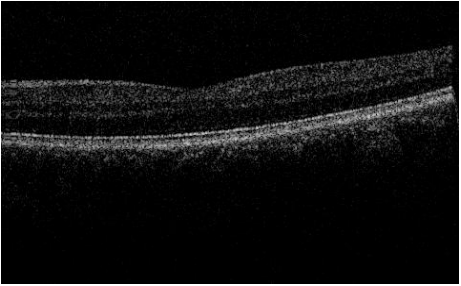

Figure 2. Speckle noises in retina OCT image

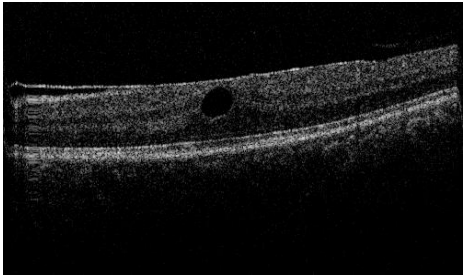

Figure 3. Abnormalities in retinal OCT images

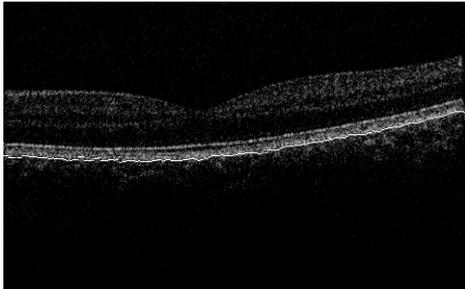

Figure 4. Discrete boundary points obtained after step 4

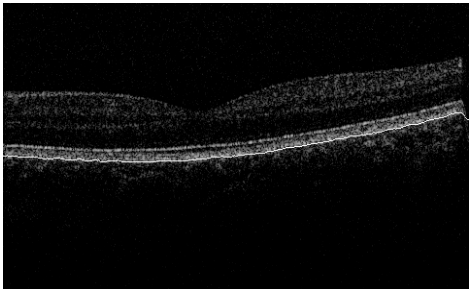

Figure 5. Smoothed boundary obtained after step 5

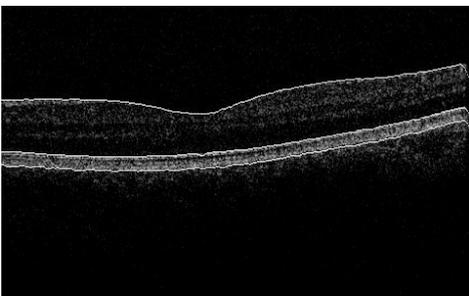

Figure 6. Three boundaries determined using our method

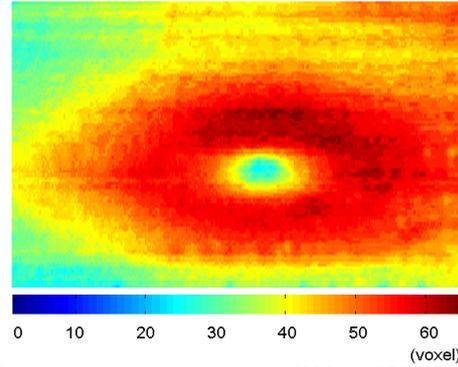

Figure 7. Retinal thickness map produced by ILM and RPE positions

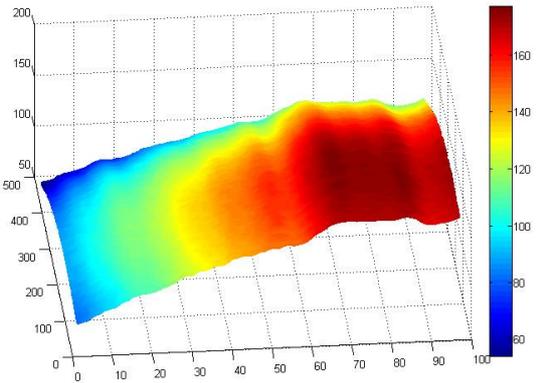

Figure 8. 3D visualization of RPE boundary surface

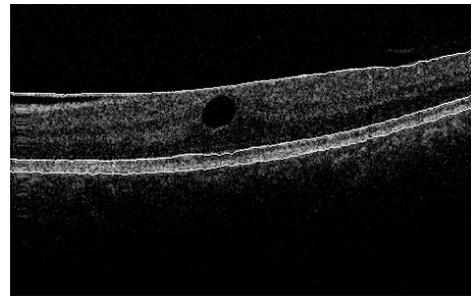

Figure 9. Three boundaries determined by our methods on abnormal retinal OCT volume data

The experiments show that our algorithm has the advantages of simplicity, efficiency and reliability, and ability to adapt to the abnormal data.

## CONCLUSION

We propose a novel 3D segmentation method of retinal OCT volume data by using pixel intensity, boundary position information, intensity changes on both sides of the border simultaneously. The method designs a specific 3D difference operator for the processing boundary to enhance the border, it conducts a three-dimensional smoothing procedure to denoise the volume data, and it further utilizes the boundary position to get a new volume data as a better foundation to find the desired boundaries. Our method can segment three important boundary surfaces at present. Experiments show that it is automatic, efficient, accurate and robust.

ACKNOWLEDGMENT

The authors would like to thank Shenzhen MOPTIM Imaging Technique Co., Ltd. for providing us with important clinical retinal OCT data which is the foundation of our work.